\documentclass[letterpaper, 10 pt, conference]{ieeeconf}
\IEEEoverridecommandlockouts
\usepackage{amsmath}
\usepackage{amsfonts}
\usepackage{bbm}
\usepackage{bbold}
\usepackage{mathtools}
\usepackage{cite}
\usepackage[table]{xcolor}
\usepackage{graphicx}
\usepackage{array,booktabs,arydshln}
\usepackage{hyperref}
\usepackage{algorithm}
\usepackage{algpseudocode}
\usepackage{caption}

\title{\LARGE \bf
Learning Perceptual Hallucination for \\Multi-Robot Navigation in Narrow Hallways
}

\author{Jin-Soo Park$^{1}$, Xuesu Xiao$^{2, 3}$, Garrett Warnell$^{4, 5}$, Harel Yedidsion$^{4}$, and Peter Stone$^{4, 6}$
\thanks{$^{1}$Department of Electrical and Computer Engineering, The University of Texas at Austin $^{2}$Department of Computer Science, George Mason University $^{3}$ Everyday Robots $^{4}$Department of Computer Science, The University of Texas at Austin $^{5}$Army Research Laboratory $^{6}$Sony AI \scriptsize\texttt{js.park@utexas.edu, xiao@gmu.edu, garrett.a.warnell.civ@army.mil, \{harel, pstone\}@cs.utexas.edu}}
}

\begin{document}

\maketitle
\thispagestyle{empty}
\pagestyle{empty}

\begin{abstract}

While current systems for autonomous robot navigation can produce safe and efficient motion plans in static environments, they usually generate suboptimal behaviors when multiple robots must navigate together in confined spaces.
For example, when two robots meet each other in a narrow hallway, they may either turn around to find an alternative route or collide with each other. 
This paper presents a new approach to navigation that allows two robots to pass each other in a narrow hallway without colliding, stopping, or waiting. 
Our approach, Perceptual Hallucination for Hallway Passing (\textsc{phhp}), learns to synthetically generate virtual obstacles (i.e., {\em perceptual hallucination}) to facilitate passing in narrow hallways by multiple robots that utilize otherwise standard autonomous navigation systems. 
Our experiments on physical robots in a variety of hallways show improved performance compared to multiple baselines.

\end{abstract}
\section{INTRODUCTION}
One of the grand goals of the robotics community is to safely and reliably deploy fully-autonomous mobile robots in common environments over extended periods of time.
Indeed, many researchers have moved toward this vision and reported hundreds of hours of unsupervised, collision-free navigation by a single robot\cite{biswas20161, khandelwal2017bwibots}.

However, long-term deployment of \emph{multiple} autonomous robots in common spaces still remains a difficult task. 
One reason for this difficulty is that, while conventional navigation systems are good at handling static environments, their performance deteriorates in the presence of dynamic obstacles, e.g., other moving robots.
The research community has explored some solutions to this problem, but these solutions typically rely on strict requirements such as a perfectly-controlled space (e.g., a warehouse) or perfect sensing \cite{van2011reciprocal}, and they cannot guarantee safety in novel environments without employing time-consuming movement schemes such as one robot halting while another moves past \cite{CoRL20-Park}.
To the best of our knowledge, there are no reports that claim long-term deployment of \emph{multiple} autonomous robots in uncontrolled spaces without human supervision.

Separately, recent work in the navigation community leveraging the concept of \emph{perceptual hallucination}\cite{xiao2021toward, xiao2021agile, wang2021agile} has demonstrated impressive results in allowing robots to navigate highly constrained spaces successfully.
Here, perceptual hallucination refers to the technique of forcing the robot to perceive specific additional virtual obstacles such that motion plans generated and executed in the presence of these additional obstacles will exhibit certain desired behaviors.
One intuitive motivation for such techniques is that the additional obstacles serve as a kind of blinder for the robot by concealing unnecessary (or even distracting) information.
To date, however, perceptual hallucination has not been applied in the context of multiple robots or dynamic obstacles.

In this paper, we hypothesize that perceptual hallucination can be used to improve conventional navigation systems in multi-robot and confined settings.
In particular, we posit that, by using perceptual hallucination techniques to obscure the presence of moving objects that would otherwise cause these conventional systems to generate suboptimal behavior (e.g., collision or turning around), we can enable multi-robot navigation in confined spaces such as narrow hallways.
If true, this would imply that hallucination would allow system designers to solve the multi-robot navigation problem using the same conventional navigation systems that have been thoroughly tested to be stable in static environments.

To investigate our hypothesis, we introduce and evaluate \emph{Perceptual Hallucination for Hallway Passing} (\textsc{phhp}), a hallucination-based approach to improve a given navigation policy in the setting of multi-robot navigation in narrow hallways.
\textsc{phhp} uses experience gathered in domain-randomized simulation episodes of hallway passing in order to learn the proper size and placement of virtual obstacles so as to enable successful navigation.
We investigate the performance and robustness of using \textsc{phhp} in common hallways with both simulation and real-world experiments, and we find that it can achieve similar performance compared to a leading existing method, Optimal Reciprocal Collision Avoidance (\textsc{orca}) \cite{van2011reciprocal}, while relaxing its assumption of perfect information about the surroundings.
We further show that, compared to a rule-based right-lane-following method, \textsc{phhp} reduces the average delay by 59.41\%.
Finally, we show that \textsc{phhp} is robust to the sim-to-real gap, different speeds, detection ranges, and even various hallway shapes and widths.

\section{RELATED WORK}
\label{sec::related}
The PHHP approach we present is a machine-learning-based solution to the problem of autonomous multi-robot navigation.
Therefore, we review here both conventional approaches that have been proposed to solve the multi-robot navigation problem, and also more recent approaches that have specifically incorporated the use of machine learning.
We also briefly review recent work on the use of perceptual hallucination in navigation.

\begin{figure*}[t]
    \centering
    \includegraphics[width=\textwidth]{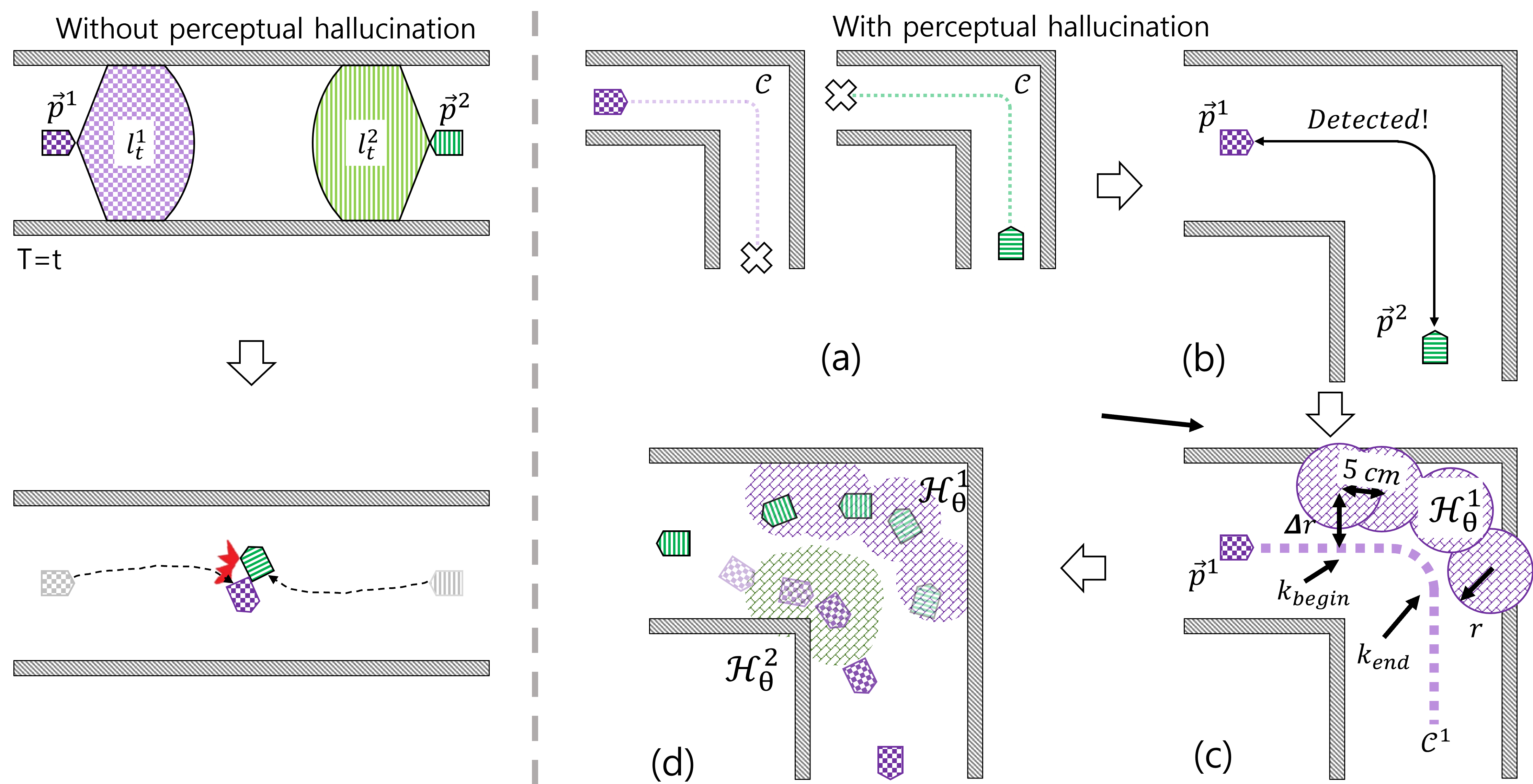}
    \caption{
    Overview of the proposed Perceptual Hallucination for Hallway Passing approach. (Left) multi-robot hallway passing scenario with an existing navigation system, and (right) how \textsc{phhp} improves the navigation system with hallucinated sensor readings.
    \textsc{phhp} deployment proceeds as follows: (a) two robots try to pass each other in the hallway; (b) one robot detects the other and initiates the hallucination process; (c) the trained \textsc{phhp} system generates a hallucinated field $\mathcal{H}_{\boldsymbol{\theta}}$ in the global coordinate system based on the global plan (d) each robot uses its corresponding hallucinated scan $\mathbf{l}_{\mathcal{H}_\theta}$ and its existing navigation system to handle the rest of the scenario.
    Note that the hallucination function, $h(\mathbf{l}, \mathcal{H})$, computes the depth value of hallucinated readings by taking the \emph{minimum} value between the real and virtual scans.
    }
    \label{fig:overview}
\end{figure*}

\subsection{Conventional Approaches}
We divide conventional approaches to multi-robot navigation into ones based on coordination between robots and approaches that operate in a fully decentralized manner.
Approaches that require tight coordination include the centralized approach deployed by Kiva Systems (now Amazon Robotics) in controlled warehouse spaces \cite{wurman2008coordinating} and the coordinated task planning approach proposed by Jiang et al. \cite{AURO19-jiang}.
While these approaches can be used successfully, the reliance on coordination limits their use in scenarios without sufficient communication bandwidth and environments with external dynamic obstacles such as third-party robots.
Conventional decentralized approaches \cite{fiorini1998motion}, including \textsc{orca} \cite{van2011reciprocal}, on the other hand, do not require explicit coordination and communication between agents, and have instead focused on modifying single-agent motion planning in an attempt to make them applicable in multi-robot settings.
However, these approaches have their own drawbacks, typically exhibiting high sensitivities to errors in dynamic obstacle state estimation, sensor noise, and parameter selection \cite{hennes2012multi, alonso2013optimal}.

\subsection{Learning-Based Approaches}
Inspired by recent success in machine learning, several in the community have proposed methods that use learning to enable multi-robot navigation \cite{xiao2022motion, mirsky2021prevention}.
Decentralized end-to-end approaches, i.e., approaches that learn mappings directly from sensor readings to motion commands, have been shown to be successful in limited settings \cite{long2017deep, long2018towards, lin2019end, tan2020deepmnavigate}, but they are typically less successful in novel environments and often suffer from a lack of safety guarantees, e.g., they may not prevent collisions in highly constrained narrow hallways.
There have also been some hybrid attempts to combine conventional navigation with machine learning for multi-robot navigation \cite{fan2020distributed,CoRL20-Park}, but they have thus far typically resulted in sub-optimal passing behaviors.

\subsection{Perceptual Hallucination}
Finally, the concept of perceptual hallucination, which our method also uses, has recently emerged as an effective tool for enabling navigation in highly-constrained spaces \cite{xiao2021toward, xiao2021agile, wang2021agile} by allowing robots to synthetically modify their own sensor readings for better neural network training, and simplified motion planning during deployment, or both.
Despite its success, however, the idea of hallucination has not previously been applied to dynamic scenarios, including the multi-robot scenario that we study here.
\section{APPROACH}
\label{sec::approach}
In this section, we first formulate the multi-robot hallway passing problem. We then describe our solution, Perceptual Hallucination for Hallway Passing (\textsc{phhp}). 

\subsection{Problem Formulation}
We consider a hallway passing scenario in which two robots must pass each other in confined hallways that are barely wide enough to permit collision-free passing.
We specifically consider this scenario in four widely observed hallway shapes: I-, L-, T-, and Z-shaped.
In this context, let $\vec{p}^1$ and $\vec{p}^2$ denote the two-dimensional global positions of the first and second robot, respectively.
Further, let $\mathcal{C}$ denote a set of global points covering the center of the hallway.
We assume that each robot is equipped with a two-dimensional LiDAR scanner, and we denote the LiDAR measurements obtained by each robot at time $t$ as $\mathbf{l}^1_t$ and $\mathbf{l}^2_t$.
We also assume that both robots use a global planner and a collision-free local navigation system designed for static environments (e.g., ROS \texttt{move\textunderscore base}\cite{ros_move_base}).

In the context of the above scenario, we seek to investigate whether perceptual hallucination can improve the performance of an existing navigation system to reduce collisions and decrease passing delay.
Mathematically, we use $h$ to denote the \emph{hallucination function}, i.e., the sensor reading $\mathbf{l}_\mathcal{H} = h(\mathbf{l}, \mathcal{H})$ is modified by transforming a LiDAR scan $\mathbf{l}$ such that it appears as if virtual obstacles specified by an {\em obstacle field} $\mathcal{H}$ (i.e., the shape and location of hallucinated obstacles) were added to the current environment.
To assure safety, $\mathbf{l}_\mathcal{H}$ only contains \emph{additional} obstacles, i.e., to compute the depth value at any particular bearing $k$, the \emph{minimum} value between the real scan, $\mathbf{l}$, and a hallucinated scan corresponding to only obstacles in $\mathcal{H}$, $\mathbf{v}_\mathcal{H}$, is sent to the robot.
Finally, we only consider here cases in which the hallucinated obstacle field is static.

In order to use perceptual hallucination, we must determine what each robot should use for its hallucinated obstacle field $\mathcal{H}$ to enable better passing.
In general, $\mathcal{H}$ could consist of an arbitrary number of obstacles, each with an arbitrary shape.
However, to make the problem tractable, we consider only obstacle fields comprised of the union of a set of contiguous circles of some fixed radius.
We denote such obstacle fields as $\mathcal{H}_{\boldsymbol{\theta}}(\mathcal{C})$ parameterized by $\boldsymbol{\theta} = (r, {\Delta}r, k_{begin}, k_{end})$, where $r$ specifies the radius of each circle,
and $k_{begin}$ and $k_{end}$ specify the fraction that divides the distance from the robot to the first and last circle along the hallway center points $\mathcal{C}$ by the detection range of the robot.
Between these starting and ending circles, $\mathcal{H}_{\boldsymbol{\theta}}(\mathcal{C})$ contains a new circle every 5cm.
Finally, $\Delta r$ specifies the distance from the center of the circle to the hallway center points, $\mathcal{C}$.
\textsc{phhp} works best with each robot using the same $\mathcal{H}_{\boldsymbol{\theta}}$, but the system can still improve multi-robot navigation ability with different $\mathcal{H}_{\boldsymbol{\theta}}$ as long as $\mathcal{H}_{\boldsymbol{\theta}}$ is valid. 
See Section \ref{ssec:robustness} for more detail.
An overview of the hallway passing scenario and how perceptual hallucination is applied is illustrated in Figure \ref{fig:overview}.

In order to quantify which $\mathcal{H}_{\boldsymbol{\theta}}$ is best for hallway passing, we define a hallway-passing cost function.
For a given hallway passing episode, we define this cost function to encourage both fast and safe passing, i.e.,
\begin{equation}
    C(\mathcal{H}_{\boldsymbol{\theta}}) = \frac{\textsc{ttd}_{1}(\mathcal{H}_\theta) + \textsc{ttd}_{2}(\mathcal{H}_\theta)}{2} + c_{coll}*\mathbb{1}_{coll} \; ,
    \label{eq:cost}
\end{equation}
where $\textsc{ttd}_{i}(\mathcal{H}_\theta)$ denotes the amount of time (seconds) it takes for robot $i$ to reach its goal using $\mathcal{H}_\theta$, and $\mathbb{1}_{coll}$ is an indicator function that is 1 if a collision occurred in the episode and 0 otherwise.
In our implementation, we set the collision penalty $c_{coll}$ to 100.
With this setup, the problem of finding the best obstacle field to hallucinate for the hallway passing problem becomes one of finding the $\boldsymbol{\theta}$ that minimizes this cost, i.e.,
\begin{equation}
    \boldsymbol{\theta}^* = \arg \min_{\boldsymbol{\theta}} C(\mathcal{H}_{\boldsymbol{\theta}}) \; .
    \label{eq:phhpprogram}
\end{equation}

\subsection{Optimal Hallucination}
\label{sec::optimalhallucination}
We solve Equation (\ref{eq:phhpprogram}) and find an effective $\boldsymbol{\theta}^*$ using the Covariance Matrix Adaptation Evolution Strategy (\textsc{cma-es}) \cite{hansen2003ecj} algorithm, a population-based, black-box optimizer that selects and evaluates successive generations of samples.
In each generation, \textsc{cma-es} samples N data points $\boldsymbol{\Theta}$ from the multivariate normal distribution and measures the cost of each sample.
Then, the mean of the next generation is updated by a weighted sum of samples in which samples with lower cost have higher coefficients. Finally, the covariance matrix is updated with the same coefficient used to update the mean with the decaying factor.
This process is used to selectively sample new generations until the covariance of the next generation is less than a threshold.
Finally, the minimum-cost sample across all generations is returned as $\boldsymbol{\theta}^*$.

To evaluate a particular sample $\boldsymbol{\theta}$ when running \textsc{cma-es}, we compute $C(\mathcal{H}_{\boldsymbol{\theta}})$ by executing a hallway passing episode with perceptual hallucination in simulation.
For each episode, two robots are initialized at each side of an I- or L-shaped hallway and given navigation goals that require them to pass one another.
Then each robot begins navigating using its base navigation system.
When robots detect one another by communicating, each robot employs \textsc{phhp} with $\mathcal{H}_{\boldsymbol{\theta}}(\mathcal{C})$ where the hallway center points $\mathcal{C}$ are approximated by the path provided by a global planner that seeks to maximize the margin of the path.
The episode ends when both robots have successfully reached their respective goal locations.
The amount of time it takes robot $i$ to reach its goal is recorded as $\textsc{ttd}_{i}$.
Collisions are defined as any time when a robot is in contact with either another robot or a wall.

In order to ensure $\boldsymbol{\theta}^*$ is robust to differences between conditions in simulation and those in the real world, we further employ \emph{domain randomization} \cite{jakobi1997, tobin2017domain}.
That is, we compute the \textsc{cma-es} objective for each sample by averaging costs obtained over several simulation episodes, each with randomized starting delay $t_i$ and detection range $D_i$ sampled from uniform random distributions, $\mathcal{U}_{[0,t_{max}]}$ and $\mathcal{U}_{[d_{min},d_{max}]}$.

The pseudocode of \emph{perceptual hallucination for hallway passing} (\textsc{phhp}) is given in Algorithm \ref{alg:train}.

\begin{algorithm}[ht]
    \begin{algorithmic}[1]
    \caption{ Find optimal Hallucination with \textsc{cma-es} }
    \label{alg:train}
    \Require $r, \Delta r, k_{begin}, k_{end}$
    \State CMAES.initialize($r, \Delta r, k_{begin}, k_{end}$)
    \State $\sigma$ $\gets$ 0.1 
    \State best\_cost $\gets \infty$
    \State ${\boldsymbol{\theta}^*} \gets$ None
    \While{$\sigma \geq threshold$}
    \State $\boldsymbol{\Theta}$ $\gets$ CMAES.generate\_samples()
    \State cost $\gets$ []
    \For{$\mathbf{k} \gets 1$ to $N$}
    \State $\boldsymbol{\theta}$ $\gets$ $\boldsymbol{\Theta}[k]$
    \State $t_{1},t_{2} \gets \mathcal{U}_{[0,t_{max}]},\mathcal{U}_{[0,t_{max}]}$
    \State $D_{1},D_{2} \gets \mathcal{U}_{[d_{min},d_{max}]},\mathcal{U}_{[d_{min},d_{max}]}$
    \State $\textsc{ttd}_{1}$, $\textsc{ttd}_{2}$, coll $\gets$ episode($\boldsymbol{\theta}$, $t_1, D_1, t_2, D_2$)
    \State cost[k] $\gets$ $\frac{\textsc{ttd}_{1} + \textsc{ttd}_{2}}{2}$ + $100\cdot\mathbb{1}_{coll}$
    \If best\_cost $\geq$ $\min$(cost) 
        \State best\_cost $\gets$ $\min$(cost)
        \State ${\boldsymbol{\theta}^*} \gets$ $\boldsymbol{\theta}$
    \EndIf
    \EndFor
    \State CMAES.optimize($\boldsymbol{\Theta}$, cost)
    \State $\sigma \gets$ CMAES.evaluate()
    \EndWhile
    \State \Return ${\boldsymbol{\theta}^*}$
    \end{algorithmic}
    
\end{algorithm}

\section{EXPERIMENTS}
\label{sec::experiments}
In order to evaluate the efficacy of Perceptual Hallucination for Hallway Passing (\textsc{phhp}), we measure:
\emph{(1)} the performance of \textsc{phhp} in different hallway shapes,
\emph{(2)} performance of \textsc{phhp} and alternative approaches with communication noise,
and \emph{(3)} whether \textsc{phhp} is robust enough to overcome the sim-to-real gap; as well as to deal with different hallway shapes, different characteristics of robots, and heterogeneous obstacle fields.

We compare \textsc{phhp} to three alternative approaches: a rule-based, right-lane-following baseline; \textsc{nh\_orca}\cite{van2011reciprocal, alonso2013optimal}; and 
the \emph{\texttt{halting} method} \cite{CoRL20-Park}.\footnote{Referred to as the {\em adaptive method} in \cite{CoRL20-Park}.}
Details of each of these approaches is provided in Section \ref{ssec:alternative}.

We evaluate the performance of each method using the following metrics:
\begin{itemize}
    \item $\Delta t$: The amount of delay compared to a single robot traversing the same hallway.
    \item $P_{collision}$: The probability of collision.
    \item $P_{failure}$: The probability that the navigation system fails to generate a valid passing plan, which typically manifests as the robot turning around.
\end{itemize}

\subsection{Platform}
We evaluate \textsc{phhp} using BWIBots\cite{IJRR17-khandelwal}, a custom differential-drive robot atop a Segway base.
A single BWIBot is 65cm wide and has a maximum linear velocity of 1.0 m/s.
The BWIBot is equipped with a front-facing Hokuyo LiDAR sensor with a 170-degree field of view and a maximum range of 20m.
For the underlying navigation system, the BWIBot uses the E-Band planner\cite{quinlan1993elastic} as the local planner, which continually generates a sequence of motion commands that result from planning over a 4m horizon.

\subsection{Training}
We train \textsc{phhp} in the widely-used Gazebo \cite{koenig2004design} simulator since it provides a safe and fast way to collect realistic data.
Two types of hallway are used in training: {\em (a)} an ``I-shaped'' straight hallway, and {\em (b)} an ``L-shaped'' hallway corner.
Both hallways are 1.6m-wide, a width for which the two 65cm-wide BWIbots can barely pass each other without colliding.
Training episodes proceed as described in Section \ref{sec::optimalhallucination}, where the robots spawn at each side of the hallway, 14m apart from one another.
As described in Section \ref{sec::approach}, domain randomization is used during simulation training in an effort to ensure that the learned policy works well when deployed in the real world.
Specifically, we impose a random starting delay from 0 to 2 seconds for each robot and we randomly set the detection range to a distance between 7 and 9 meters.

We use \textsc{phhp} to find virtual obstacles for each type of hallway; I-shaped and L-shaped.
To accelerate the \textsc{cma-es} search, we used $(r, \Delta r, k_{begin}, k_{end}) = (0.5, 0.05, 0.3, 0.6)$ as an initial hypothesis, which, intuitively, represents a virtual obstacle that entirely blocks the left half of the hallway.
For each run of \textsc{cma-es}, a total of approximately 100 generations occur before the standard deviation of all samples in a generation becomes less than our selected threshold of 0.01.
A single generation contains 8 sample configurations, and each configuration is evaluated by the average cost in Equation (\ref{eq:cost}) averaged over 200 domain-randomized episodes.
Training takes about 20 hours using a distributed computing cluster with about 150 nodes.
The identified configurations of virtual obstacle fields $\mathcal{H}_{\boldsymbol{\theta}^*}$ in two types of hallways are presented in Table \ref{tbl:cmaes_results}. 

\begin{table}[ht]
  \centering
  \caption{Learned configuration of $\mathcal{H}_{\boldsymbol{\theta^*}}$ in various hallways.}
  \begin{tabular}{|c|cccc|}
    \hline
    train environment & radius & $\Delta r$ & $k_{begin}$    & $k_{end}$    \\ \hline
    I-shape hallway   & 0.7590  & 0.7888 & 0.4845 & 0.4910 \\
    L-shape hallway   & 0.5122  & 0.5661 & 0.4842 & 0.5001 \\
    \hline
  \end{tabular}
  \label{tbl:cmaes_results}
\end{table}

\subsection{Alternative Approaches} \label{ssec:alternative}
We compare \textsc{phhp} with three alternative methods; a right-lane-following baseline, \textsc{nh\_orca}, and the halting method.

The right-lane-following baseline, or simply \texttt{baseline} is inspired by the US traffic standard.
It is a rule-based algorithm that, upon detection of the other robot, moves the robot into a human-annotated right lane and proceeds in that lane until the two robots pass one another.
However, the baseline has two drawbacks; (a) it requires human effort to manually specify the lanes for each hallway in the environment, and (b) since the robot needs to always stay in a narrow lane even when another robot is not present, the average speed of the robot drops significantly.

\textsc{nh\_orca}\cite{alonso2013optimal} is an extension of \textsc{orca}\cite{van2011reciprocal} that can be applied to differential drive robots by approximating the trajectory of \textsc{orca} with differential drive constraints.
\textsc{orca} finds the optimal collision-avoiding velocity command with minimum deviation from the velocity command that heads directly to the goal.
As long as \textsc{orca} obtains accurate information about the environment with sufficient frequency (i.e., the precise position and velocity of the other robot), it is able to provide collision avoidance behavior with a small delay.
However, while \textsc{orca} provides excellent performance (we consider it to be an upper bound) in simulation with perfect communication channels, real robots often must rely on noisy communication channels to share position and velocity information, which we will show degrades the performance of \textsc{orca}.
\textsc{phhp}, on the other hand, only needs to observe the presence of the other robot once.

Finally, the \texttt{halting} method\cite{CoRL20-Park} is a system designed for hallway passing in which, when a halting robot detects a potential collision, the halting robot immediately moves to the nearest, safe parking spot until the non-halting robot completely passes and then resumes.
While this approach can be used to avoid collisions in some hallways settings, the halting behavior itself typically causes the average delay to be high.

\subsection{Testing \textsc{phhp} in Different Hallway Shapes}
\label{sec::hallwayshapes}
\begin{figure}
    \centering
    \includegraphics[width=0.48\textwidth]{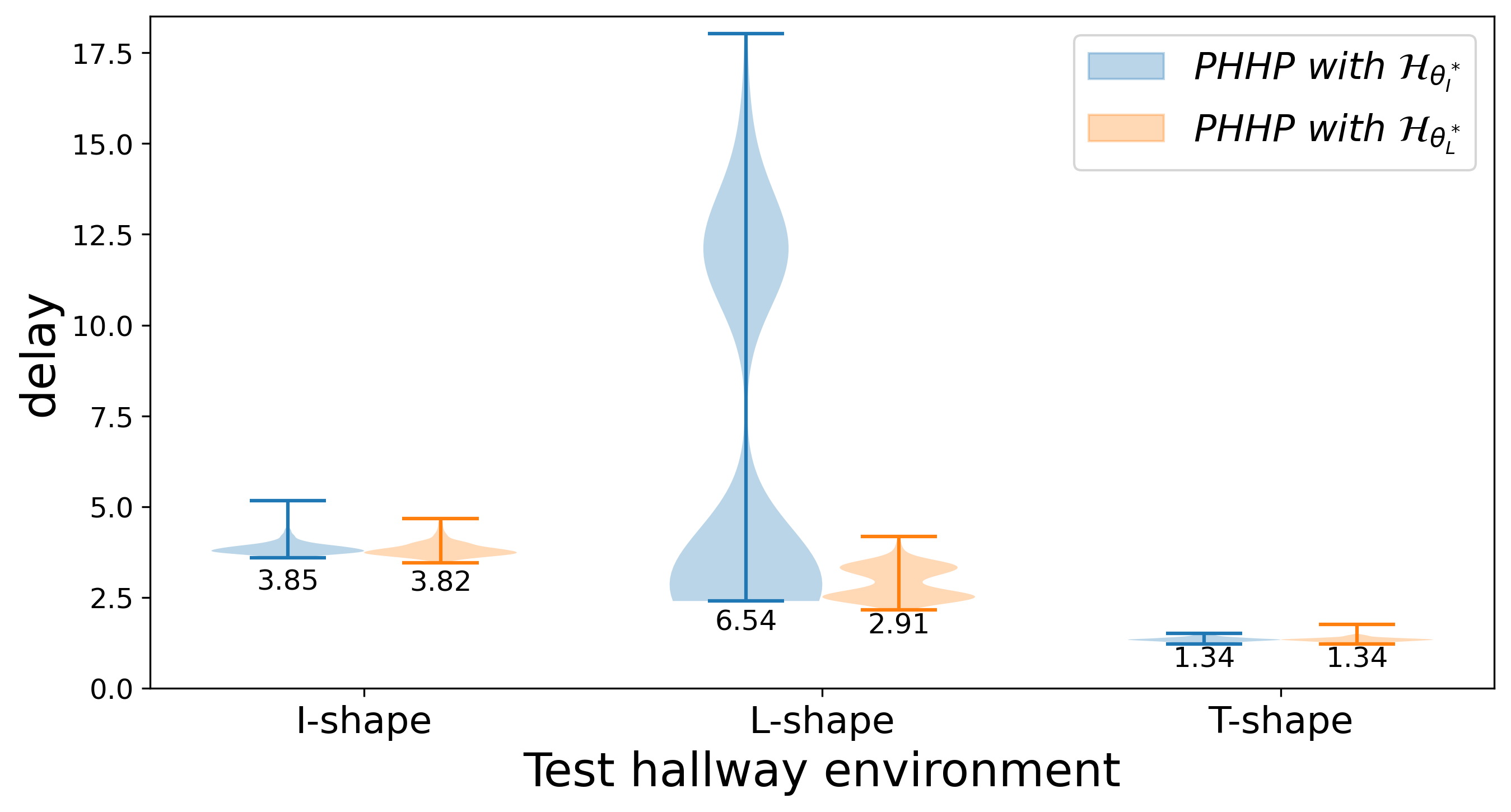}
    \caption{
    The figure shows the violin plot of \textsc{phhp} trained in the subscripted environment deployed in various hallways.
    Each bar consists of 300 test episodes in the given hallway in simulation.
    Both \textsc{phhp} with $\mathcal{H}_{\boldsymbol{\theta^*_{I}}}$ and $\mathcal{H}_{\boldsymbol{\theta^*_{L}}}$ provide smooth interaction in the I and T-shaped hallway.
    However, $\mathcal{H}_{\boldsymbol{\theta^*_{I}}}$ often fails to provide a smooth solution in the L-shaped hallway while $\mathcal{H}_{\boldsymbol{\theta^*_{L}}}$ does.
    Note that no collision happened over 1,800 episodes in simulation.
    The obstacle field used by each \textsc{phhp} can be found in Table \ref{tbl:cmaes_results}.
    }
    \label{fig:performance_between_phhp}
\end{figure}

We trained $\textsc{phhp}$ in both I- and L-shaped hallways, and tested the resulting obstacle fields, $\mathcal{H}_{\boldsymbol{\theta^*_I}}$ and $\mathcal{H}_{\boldsymbol{\theta^*_L}}$, 300 times each in I-, L-, and T-shaped hallways.
Note that the T-shaped hallway is a test hallway that has never been exposed to the robot during training.
The results can be found in Figure \ref{fig:performance_between_phhp}, where we can see that the performance of \textsc{phhp} using $\mathcal{H}_{\boldsymbol{\theta^*_L}}$ outperforms $\mathcal{H}_{\boldsymbol{\theta^*_I}}$ in all three environments, presumably because the wings of the L-shaped hallway can be viewed as an I-shaped hallway with a shorter length.
For this reason, we use \textsc{phhp} with $\mathcal{H}_{\boldsymbol{\theta}^*_{L}}$ to represent \textsc{phhp} in further experiments unless otherwise mentioned.

\subsection{Performance in the Presence of Communication Noise}
\begin{figure}[t]
    \centering
    \includegraphics[width=0.48\textwidth]{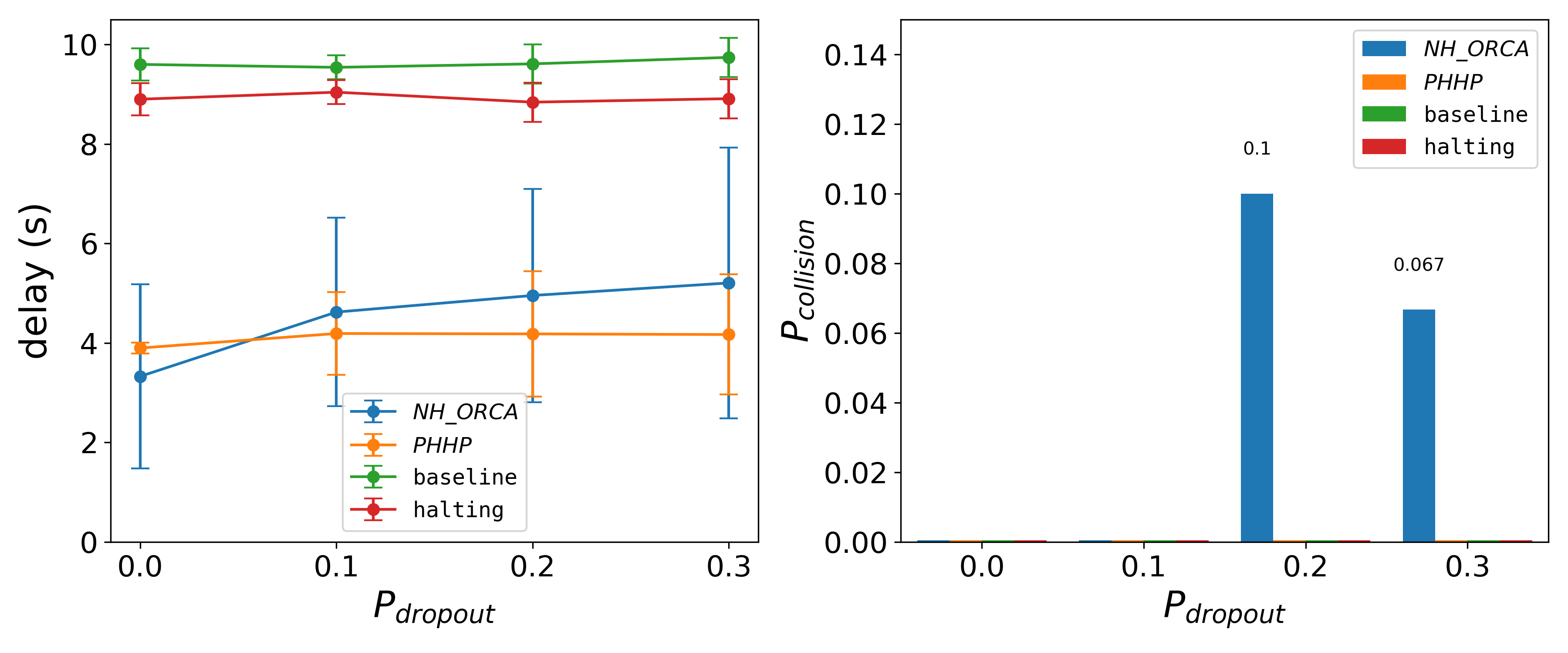}
    \caption{
    The performance analysis of each algorithm in the noisy communication channel; (left) the average delay of each algorithm, and (right) the probability of collision.
    The average delay of \textsc{phhp} remains consistently stable, while the delay of \textsc{nh\_orca} increases rapidly as the noise increases.
    The right-lane-following baseline and halting methods are also resistant to noise, but the average delay of both algorithms is much higher than \textsc{phhp}.  
    }
    \label{fig:performance}
\end{figure}
We now seek to investigate the efficacy of \textsc{phhp} versus alternative approaches.
In particular, we are interested in how robust each approach is to communication noise, and so we perform our experiments here in a setting in which an artificial noisy channel is imposed using a dropout model.
Under this model, messages fail to reach their destination with a specified probability and, when such a failure occurs, each algorithm uses the last received message to perform collision avoidance.
We expect that \textsc{phhp}, the baseline method, and the halting method, all of which only require observing the other robot once before activating the corresponding collision avoidance behavior, will slowly degrade as the probability of dropout increases.
In contrast, we expect that the performance of ORCA will drop rapidly as we increase channel noise.
Experiments were conducted 30 times in the 1.6m-wide I-shaped hallway in Gazebo, and the results are shown in Figure \ref{fig:performance}.

The results indicate that, as expected, the average delay incurred by \textsc{phhp} becomes lower than that of \textsc{nh\_orca} as channel noise increases.
Additionally, about 10\% of \textsc{nh\_orca} trials ended in collision in high noise settings, while no collisions were reported for \textsc{phhp} across all noise levels.
Taken together these results suggest that \textsc{phhp} will fare better than \textsc{nh\_orca} in real-world deployment settings.
Additionally, we note that, while the \texttt{baseline} and \texttt{halting} methods are robust to noise, the raw performance of those methods is far worse than that of \textsc{phhp}.

\subsection{Robustness Analysis} \label{ssec:robustness}
\begin{figure}
    \centering
    \includegraphics[width=0.53\textwidth]{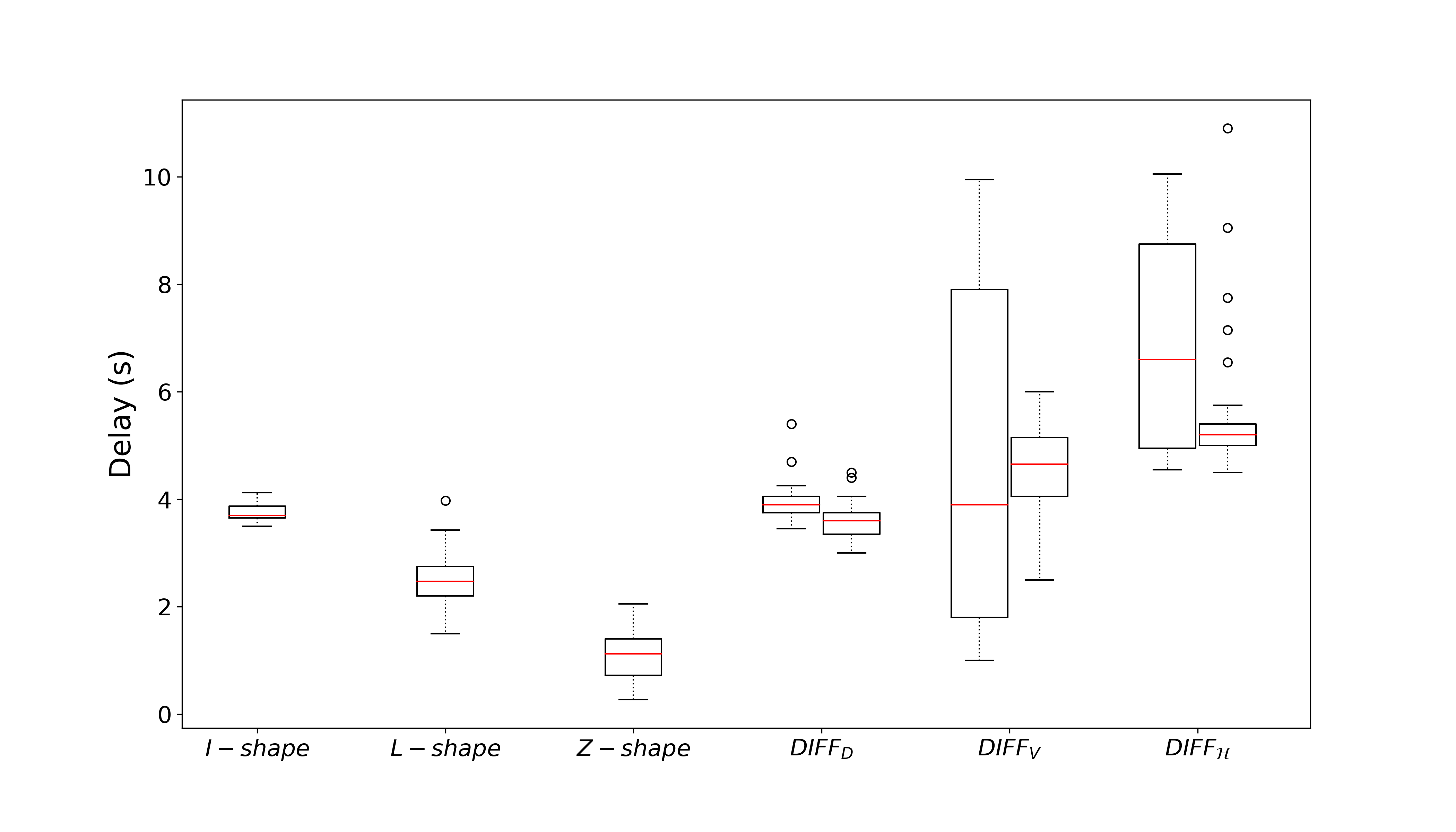}
    \caption{
    Box and whisker plot of real-world experiments with various conditions.
    Details for each condition can be found in Table \ref{tbl:exp_setups}.
    When the two robots use different conditions, the delay of each robot is plotted separately.
    Across conditions, \textsc{phhp} incurs between just 1 and 7 seconds of average delay compared to single-robot navigation in the same hallway, and did not incur a single collision or turnaround behavior across all 192 experiments represented here.
    }
    \label{fig:robustness}
\end{figure}

\begin{figure}
    \centering
    \includegraphics[width=0.40\textwidth]{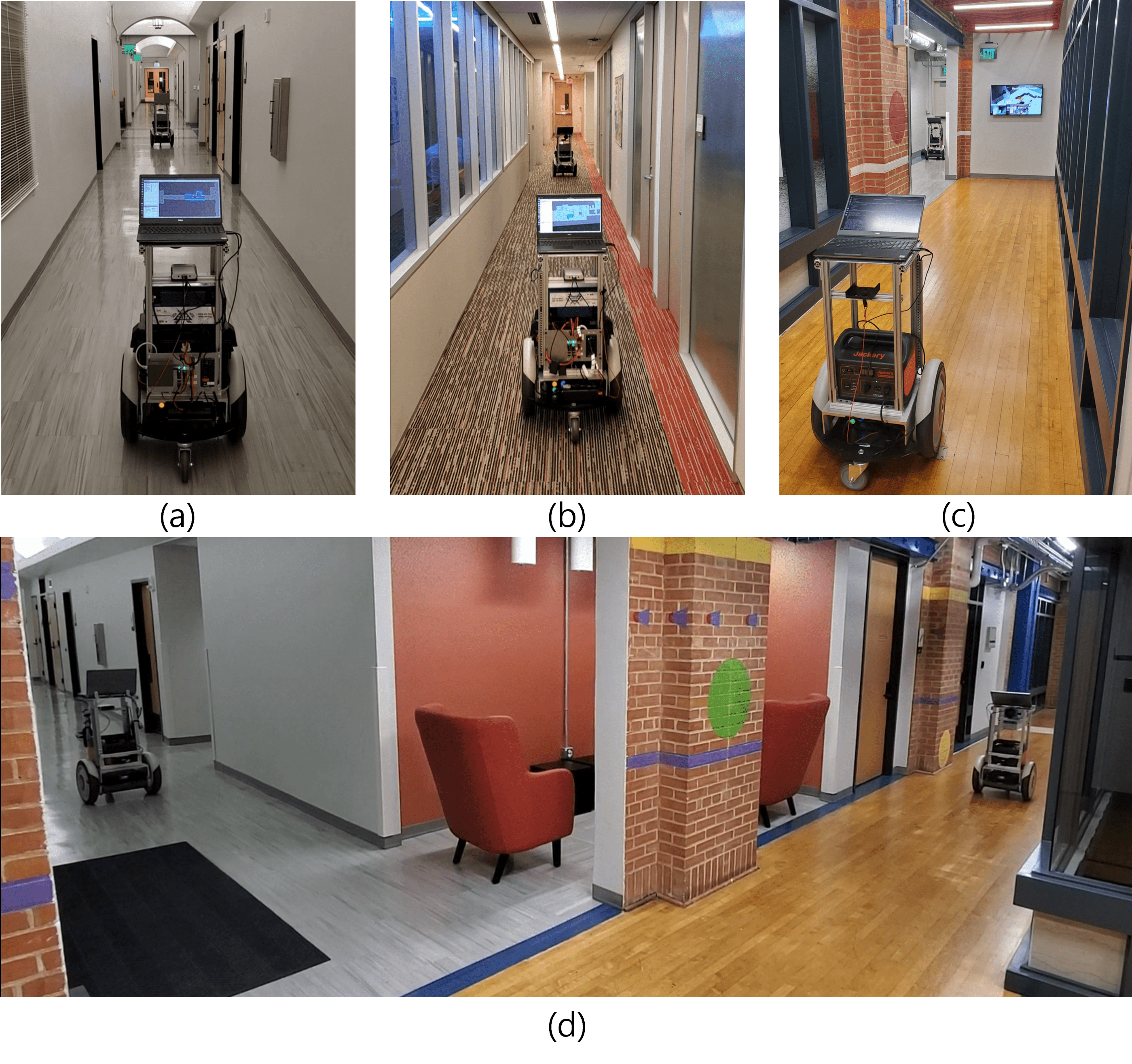}
    \caption{
    The photos of hallways in which we performed the real-world \textsc{phhp} experiments; (a) 1.8m I-shaped, (b) 1.6m I-shaped, (c) 1.6m Z-shaped, and (d) L-shaped.
    }
    \label{fig:real_hallways}
\end{figure}

We investigate the robustness of \textsc{phhp} in terms of sim-to-real transfer performance in different environments, different characteristics of the robot (i.e., detection range and velocity), and even with different but valid obstacle fields, 
$\mathcal{H}_{\boldsymbol{\theta}^*}$,
used by \textsc{phhp}.
The test setup is as follows.
We directly deploy simulation-trained \textsc{phhp} in the real world and conduct 32 evaluation episodes per condition, each with a specific environment and robot parameters as shown in Table \ref{tbl:exp_setups}.
``ID'' defines the name of each condition, while ``Hallway Type'' represents the hallway shape and width, where L-shaped hallways are 1.8m wide on one segment and 1.6m wide on the other.
Example hallways can be seen in Figure \ref{fig:real_hallways}.
$\texttt{D}_i$ and $\texttt{V}_i$ describe the detection range and velocity of a robot, respectively, where the subscript $i$ denotes which robot the parameter pertains to. Finally, $\mathcal{H}_i$ represents the specific obstacle field used by \textsc{phhp} during the experiment as defined in Section \ref{sec::hallwayshapes}.
The robots report their locations to each other over WiFi.

The result of each experiment is shown in Figure \ref{fig:robustness}. 
Regarding sim-to-real transfer performance, \textsc{phhp} fares well: the average delay of \textsc{phhp} in the real I- and L-type corridors is similar to the results obtained in simulation.
Interestingly, a Z-shaped hallway only records one second of delay, presumably because the interaction in a Z-shaped hallway happens on the wider side of the hallway.
Regarding robustness to different robot characteristics, even though the maximum linear velocity of the robots significantly differed, 1.0 m/s and 0.6 m/s, respectively, \textsc{phhp} managed to resolve the hallway passing problem without any turnaround or collision.
Finally, regarding robustness to different but valid obstacle fields, the \textbf{\texttt{DIFF}$_\mathcal{H}$} results indicate that, as long as individual obstacle fields can resolve the hallway passing problem, a combination of them can also be successful.
\textbf{\textit{Importantly, no collisions or turnarounds were observed}} during the entire set of 192 real-world episodes.
These results suggest that \textsc{phhp} is very robust to the sim-to-real gap, environmental changes, different speeds and detection ranges of the robot, and even to using obstacle fields trained in different hallways.

\begin{table}[]
    \centering
    \caption{The configuration used in real world experiments.}
\begin{tabular}{l|lllllll}
ID      & Hallway Type              & $\texttt{D}_1$ & $\texttt{D}_2$ & $\texttt{V}_1$ & $\texttt{V}_2$ & $\mathcal{H}_1$ & $\mathcal{H}_2$         \\ \hline
\textbf{I-shape} & \textbf{1.8m I-shaped}     & 8.0          & 8.0          & 1.0 & 1.0          & $\mathcal{H}^*_{L}$  & $\mathcal{H}^*_{L}$          \\
\textbf{L-shape} & \textbf{L-shaped} & 8.0          & 8.0          & 1.0 & 1.0          & $\mathcal{H}^*_{L}$  & $\mathcal{H}^*_{L}$          \\
\textbf{S-shape} & \textbf{1.6m Z-shaped}     & 8.0          & 8.0          & 1.0 & 1.0          & $\mathcal{H}^*_{L}$  & $\mathcal{H}^*_{L}$          \\
\textbf{$\texttt{DIFF}_D$} & 1.8m I-shaped    & \textbf{9.0} & \textbf{7.0} & 1.0 & 1.0          & $\mathcal{H}^*_{L}$  & $\mathcal{H}^*_{L}$          \\
\textbf{$\texttt{DIFF}_V$} & L-shaped & 8.0         & 8.0          & 1.0 & \textbf{0.6} & $\mathcal{H}^*_{L}$  & $\mathcal{H}^*_{L}$          \\
\textbf{$\texttt{DIFF}_\mathcal{H}$} & \textbf{1.6m I-shaped}     & 8.0          & 8.0          & 1.0 & 1.0          & $\mathcal{H}^*_{L}$  & \textbf{$\mathbf{{H}}^*_{I}$}
\end{tabular}
    \label{tbl:exp_setups}
\end{table}

\section{CONCLUSIONS AND FUTURE WORKS}
\label{sec::conclusions}
In this paper, we presented Perceptual Hallucination for Hallway Passing (\textsc{phhp}), a new method that enables multi-robot navigation in constrained spaces.
We showed how to find the best obstacle for \textsc{phhp} to hallucinate for a given environment and navigation policy using \textsc{cma-es}.
The simulation and real-world deployment results indicate that \textsc{phhp} achieves comparable performance against \textsc{orca}, while removing the assumption that the robot has continuous access to the other robot's position and velocity.
Moreover, \textsc{phhp} outperforms both a right-lane-following baseline and our prior work, the halting method, in terms of delay.
Additionally, real-world deployment results experimentally confirm that \textsc{phhp}, which is trained in simulation, can successfully be deployed in a wide variety of real-world settings, including those in which the size of the hallway is changed, the robots move with different velocities, their perception systems exhibit a different detection range, or the virtual obstacles used by \textsc{phhp} is different.
Despite the successes we presented, \textsc{phhp} has only been developed and evaluated here with two robots and we have only explored using a particular class of hallucinated obstacle fields comprised of a number of circles.
Therefore, an important direction for future work is to investigate how to expand \textsc{phhp} to work with multiple, arbitrarily shaped obstacles in a wider variety of settings. 

\section*{ACKNOWLEDGMENTS}
\small{
This work has taken place in the Learning Agents Research Group (LARG) at UT Austin.  LARG research is supported in part by NSF (CPS-1739964, IIS-1724157, FAIN-2019844), ONR (N00014-18-2243), ARO (W911NF-19-2-0333), DARPA, GM, Bosch, and UT Austin's Good Systems grand challenge.  Peter Stone serves as the Executive Director of Sony AI America and receives financial compensation for this work.  The terms of this arrangement have been reviewed and approved by the University of Texas at Austin in accordance with its policy on objectivity in research.}

\bibliographystyle{IEEEtran}
\bibliography{IEEEabrv, references}
\addtolength{\textheight}{-12cm}

\end{document}